\documentclass[journal]{IEEEtran}
\ifCLASSINFOpdf
\else
\fi
\usepackage{amsthm}

\usepackage{graphicx}

 \usepackage{multirow}
\usepackage{graphicx}
\usepackage{amsmath} 
\usepackage{tcolorbox}
\begin{document}
\title{Thermal-Only Crowd Counting with Deployment-Time Privacy Protection}

\author{Yifei Qian,
        Zhongliang Guo, 
        Chun Tong Lei,
        Bowen Deng,
        Chun Pong Lau,~\IEEEmembership{Member, ~IEEE},\newline
        Xiaopeng Hong,~\IEEEmembership{Senior Member,~IEEE},
        and~Michael P. Pound% <-this % stops a space
\thanks{Yifei Qian, Bowen Deng and Michael P. Pound are with the School of Computer Science, University of Nottingham, NG7 2RD Nottingham, U.K.\linebreak Email:\{Yifei.Qian1, Bowen.Deng, Michael.Pound\}@nottingham.ac.uk}% <-this % stops a space
\thanks{Zhongliang Guo is with the School of Computer Science, University of St Andrews, KY16 9AJ St Andrews, U.K.\linebreak Email:\{zg34\}@st-andrews.ac.uk}
\thanks{Chun Tong Lei and Chun Pong Lau are with the Department of Data Science, City University of Hong Kong, Hong Kong SAR, China (E-mail: \{ctlei2, cplau27\}@cityu.edu.uk).}
\thanks{Xiaopeng Hong is with the Harbin Institute of Technology, Harbin, Heilongjiang, 150001, China. E-mail: hongxiaopeng@ieee.org.}% <-this % stops a space
\thanks{Manuscript received April 19, 2005; revised August 26, 2015.}}


\markboth{Journal of \LaTeX\ Class Files,~Vol.~14, No.~8, August~2015}%
{Shell \MakeLowercase{\textit{et al.}}: Bare Demo of IEEEtran.cls for IEEE Journals}



\maketitle

\begin{abstract}
While RGB-Thermal crowd counting has shown promise, the paradigm faces critical limitations: RGB data raises privacy concerns in public surveillance, and multi-modal misalignment degrades fusion performance. We propose the first thermal-only framework specifically designed for privacy-conscious crowd counting, eliminating RGB dependency at inference time and substantially reducing the privacy exposure associated with continuous RGB capture in public surveillance deployments. To mitigate thermal ambiguity, we leverage depth-to-RGB diffusion models as a cross-modal bridge, extracting discriminative features that enhance thermal representations. Critically, we demonstrate that single-step LCM denoising yields features most faithful to the structural content of the depth conditioning signal, while multi-step approaches progressively decouple features from the conditioning input and accumulate errors that degrade counting accuracy. Experiments on RGBT-CC and DroneRGBT datasets show our method achieves competitive performance against state-of-the-art RGB-T fusion methods, while requiring only thermal input during inference, eliminating the need for continuous RGB capture that constitutes the primary privacy concern in real-world surveillance deployment. The code will be made publicly available.
\end{abstract}


% \fi
\IEEEpeerreviewmaketitle

\section{Introduction}

Crowd counting estimates the number of people in images or videos for applications including public safety, urban planning, and event management. Despite recent advances, it remains challenging due to high crowd density variation, occlusions, and diverse environmental conditions~\cite{huang2023counting, 10879542, 10057072}.
Conventional RGB methods struggle in adverse conditions like night-time or poor illumination, and raise privacy concerns by capturing identifiable features. RGB-Thermal (RGB-T) crowd counting~\cite{liu2021cross} addresses robustness issues by incorporating thermal imaging for enhanced performance across varying lighting conditions, though privacy concerns from RGB persist. Existing RGB-T methods are built on the consensus that RGB and thermal modalities provide complementary information: RGB offers rich appearance features in well-lit environments but suffers from illumination dependence, while thermal captures heat signatures regardless of lighting but struggles with discrimination between humans and other heat sources. Based on this consensus, researchers focus on designing effective fusion strategies to leverage both modalities~\cite{zhou2023mc, kong2024cross, mu2025misf}.

\begin{figure}[t]
    \centering
    \includegraphics[width=0.95\linewidth]{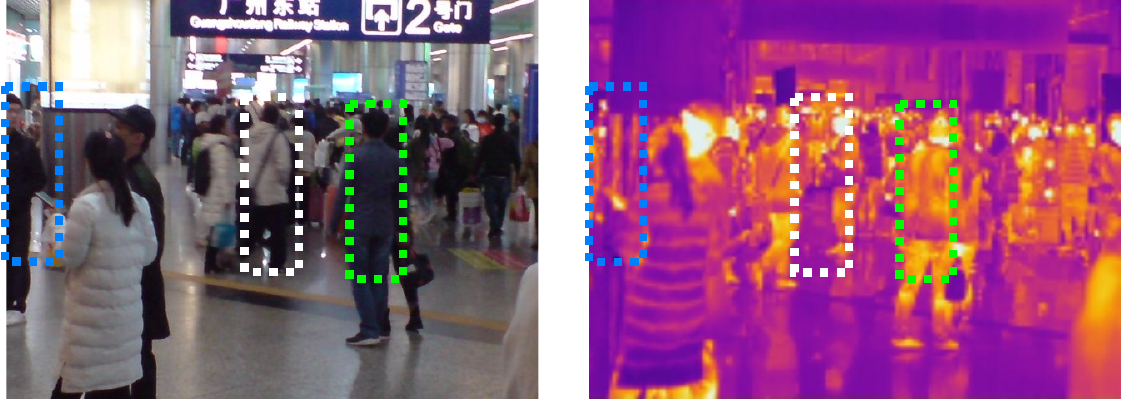}
    \caption{Misalignment issues between RGB and thermal images in RGBT datasets. Highlighted regions demonstrate spatial and temporal inconsistencies where objects appear differently across modalities.}
    \label{fig:teaser}
\end{figure}

However, as research in RGB-T crowd counting advances, we observe diminishing returns in performance gains despite increasingly sophisticated fusion strategies. An important observation is that virtually all fusion methods are implicitly built upon an idealized assumption: perfect or near-perfect alignment between RGB and thermal data. In practice, however, RGB-T pairs often exhibit spatial and temporal inconsistencies due to differences in sensor placement and acquisition timing~(Figure~\ref{fig:teaser}), introducing additional complexity that fusion strategies must explicitly or implicitly account for. While recent works have begun to address this challenge, the progress~\cite{Wang_2024_BMVC} has been incremental, suggesting that circumventing the alignment problem may be a more tractable path than solving it directly.
% In reality, this assumption rarely holds true. As shown in Figure~\ref{fig:teaser}, significant misalignment between RGB and thermal images exists due to differences in sensor setup and acquisition timing. This inconsistency between modalities introduces noise during fusion and often leads to suboptimal performance, undermining the theoretical advantages of dual-modality approaches.
These observations lead us to a complementary hypothesis: could a single well-utilized modality match complex dual-modality systems through more effective feature utilization, while avoiding the alignment challenges and privacy risks inherent to RGB-T fusion?

To explore this hypothesis, we investigate whether a carefully designed thermal-only model can achieve performance comparable to state-of-the-art RGB-T fusion methods, thereby demonstrating that the RGB modality is not indispensable for accurate crowd counting. Thermal-only approaches offer two fundamental advantages over RGB-dependent methods. First, thermal imaging maintains consistent functionality across all lighting conditions, including complete darkness where RGB cameras fail entirely. While thermal imaging inherently exhibits certain ambiguities in heat signature interpretation, these can be enhanced through complementary semantic information. This is a more tractable approach than addressing the fundamental absence of visual data in RGB streams during low-light conditions. Second, thermal imaging substantially reduces privacy exposure in public surveillance by capturing heat 
signatures rather than visually identifiable features such as faces, aligning with growing privacy awareness requirements in public monitoring applications. It is important to clarify the scope of our privacy-conscious design: we distinguish between 
training-time data usage—where RGB is used once, in a controlled setting on publicly released benchmarks, to learn cross-modal priors—and deployment-time data capture, where RGB would otherwise be continuously recorded in public spaces without consent. Our framework eliminates the latter, which constitutes the dominant privacy exposure vector in real-world surveillance. We acknowledge that thermal imaging itself retains residual privacy implications under certain conditions; our design goal is therefore risk mitigation rather than absolute privacy guarantees.

In this work, to the best of our knowledge, we propose TDCount, the first framework specifically designed for thermal-only inference in RGB-T crowd counting scenarios. To address the inherent ambiguity in thermal representations, we employ a pretrained depth-to-RGB diffusion model as the source of complementary semantic information. By extracting depth maps from thermal images and leveraging the rich priors learned by diffusion models, we enhance the discriminative capability of thermal features without requiring RGB data during inference. Rather than standard LDMs, which require hundreds of denoising steps and are computationally prohibitive as feature extractors, we adopt LCM, which enables meaningful feature extraction in just a few steps via its consistency constraint. A key design choice is to fix the initial noisy latent once and reuse it throughout training and inference, making the feature extraction process fully deterministic and trainable. Thermal representations are further aligned to the diffusion model's semantic space via text-guided prototypes. Crucially, we use only the first LCM denoising step for feature extraction, as we find additional steps progressively decouple features from the depth conditioning signal by prioritizing perceptual refinement over structural fidelity, accumulating errors that degrade counting accuracy. In summary, our main contributions are threefold:
\begin{itemize}
    \item We propose the first framework, TDCount that enables thermal-only inference within the RGB-T crowd counting paradigm by leveraging diffusion-based cross-modal priors, eliminating RGB dependency at deployment.
    \item We introduce an LCM-based feature learning framework for thermal crowd counting and reveal that the initial denoising step provides more discriminative features than multi-step denoising, which progressively accumulates errors.
    \item Our privacy-conscious design substantially reduces privacy exposure in public surveillance by eliminating continuous RGB capture at deployment time which is the primary privacy risk vector in real-world surveillance systems, while maintaining competitive accuracy compared to state-of-the-art dual-modality methods.
\end{itemize}

\section{Related Works}

\subsection{Single-Modal Crowd Counting:}
Crowd counting~\cite{qian2024semi, qian2025perspective, 10680129} has witnessed significant progress driven by deep learning advances. Early CNN-based methods focus on designing multi-scale architectures to handle scale variations in crowd scenes. MCNN~\cite{MCNN} adopts multi-column networks with different receptive fields, while CSRNet~\cite{li2018csrnet} employs dilated convolutions to capture broader contextual information. Beyond architecture design, several works address the annotation noise inherent in point-supervised crowd counting. BL~\cite{ma2019bayesian} formulates a Bayesian loss to directly learn from point annotations, and subsequent methods incorporate optimal transport theory~\cite{wang2020DMCount, Lin2021DMM} for more principled supervision. Other approaches leverage auxiliary tasks such as foreground segmentation~\cite{shi2023focus} and depth estimation~\cite{Gao2019PCCNP} to provide additional learning signals. More recently, transformer-based methods~\cite{sun2021boosting, qian2022seg, Lin_cvpr22} have been introduced to model long-range spatial dependencies, further advancing counting accuracy. Despite these advances, single-modal RGB methods remain fundamentally limited by their sensitivity to illumination changes and adverse environmental conditions, motivating the exploration of multi-modal approaches.

\subsection{RGB-T Crowd Counting:}
RGB-T crowd counting approaches emerged to address RGB methods' limitations in adverse lighting conditions.
% Early methods like BL+IADM~\cite{liu2021cross} demonstrated that even simple fusion strategies could significantly improve counting performance, establishing the consensus that RGB and thermal modalities provide complementary information and motivating subsequent research to design various sophisticated fusion strategies.
The pioneer work, BL+IADM~\cite{liu2021cross}, demonstrated that the fusion of RGB and thermal could significantly improve the counting performance, promoting the following researchers to focus on more powerful fusion strategies.
Tang et al.~\cite{tang2022tafnet} propose a three-branch fusion architecture with an information improvement module to enhance feature representations. Zhang et al.~\cite{zhang2022spatio} proposes cross-modal spatio-channel attention blocks. 
BGDFNet~\cite{xie2024bgdfnet} introduces bidirectional gating for cross-modal feature interaction and utilized dynamic convolution with multiple kernels to perceive varying pedestrian head features. Some methods also explore transformer structures~\cite{liu2023rgb, kong2024cross} to enhance the global relationship modeling for RGB-T crowd counting tasks. Meng et al.~\cite{meng2025free} propose a training enhancement strategy that introduces post-pre-training cross-modal alignment and regional density supervision without additional data or parameters.

However, the fundamental alignment challenges between RGB and thermal modalities have received limited attention. Recent work by Wang et al.~\cite{Wang_2024_BMVC} attempts to address this through a cross-modal emulation process, but the suboptimal performance caused by inherent image quality issues proves difficult to resolve through modular design alone, resulting in only marginal improvements. Moreover, the persistent dependence on RGB data in 
current RGBT approaches raises growing privacy concerns in surveillance applications, where continuous RGB capture of members of the public constitutes a primary privacy exposure vector that existing methods have yet to address.
\begin{figure*}[t]
    \centering
    \includegraphics[width=\linewidth]{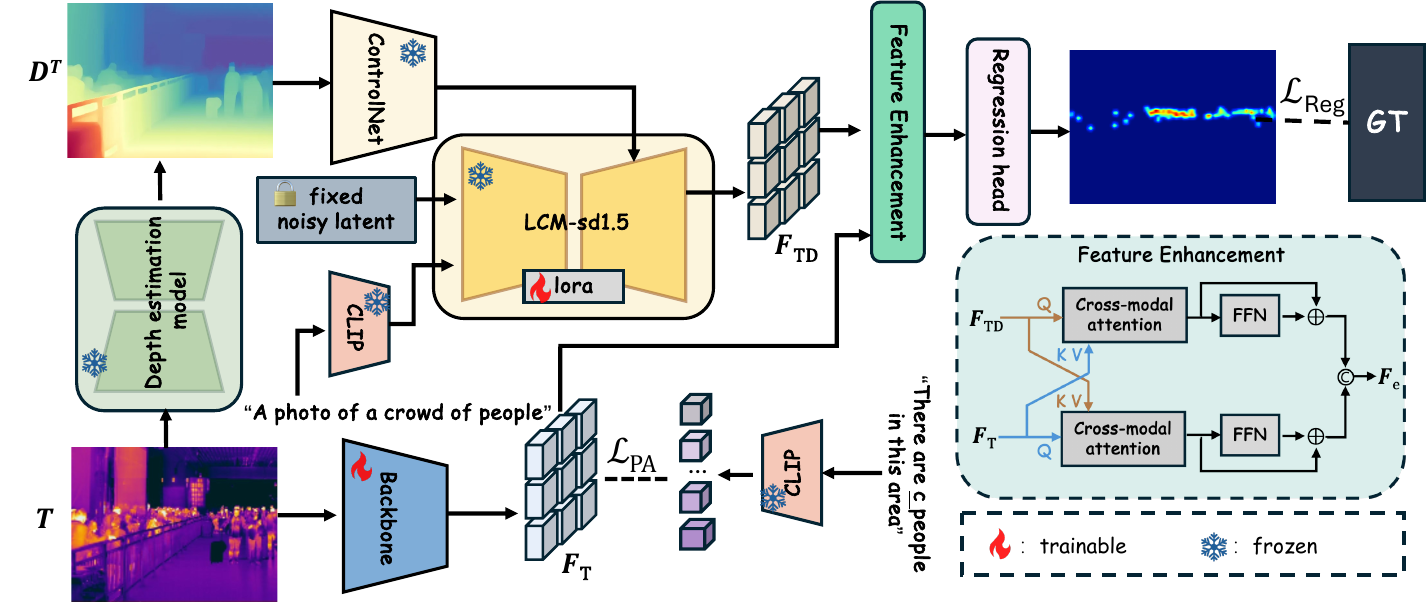}
    \caption{The overall framework of the proposed \textbf{TDCount}. We leverage ControlNet-based depth-to-RGB translation to extract complementary features $F_\text{TD}$, which are integrated with thermal features $F_\text{T}$ through a simple feature enhancement module, and aligned via a prototype alignment loss~($\mathcal{L}_\text{PA}$) for robust crowd density estimation. }
    \label{fig:overall}
\end{figure*}

\subsection{Cross-Modal Generation:}
Cross-modal generation has witnessed remarkable progress with the advent of diffusion models~\cite{NEURIPS2020_4c5bcfec}, which excel at capturing complex data distributions and generating high-quality outputs across different modalities.
UNIT-DDPM~\cite{sasaki2021unit} first achieves unpaired image-to-image~(I2I) translation using diffusion models, pioneering the approach of joint distribution modeling and Markov chain sampling. Subsequently, SDM~\cite{wang2022semantic} introduces the SPADE spatially-adaptive normalization module, efficiently embedding semantic labels into the diffusion process and achieving comprehensive superiority over GAN methods~\cite{isola2017image} in label-to-image tasks. BBDM~\cite{BBDM} realizes bidirectional mapping between domains through Brownian bridge diffusion processes, effectively alleviating the heterogeneous distribution gap. ControlNet~\cite{zhang2023adding} enables structure-guided image generation through various control signals such as depth, canny edges, pose, and semantic layouts. 
Building upon these advances, we employ a depth-to-RGB ControlNet to provide additional discriminative features for thermal representations, thereby circumventing the misalignment issues 
in RGB-T fusion. Furthermore, we employ Latent Consistency Models (LCMs)~\cite{luo2023lcm} instead of standard Stable Diffusion to reduce computational burden, and discover that single-step denoising is sufficient for effective feature extraction.

\section{Methodology}
Given paired RGB and thermal crowd images $\{(R_i, T_i)\}_{i=1}^N$, existing 
RGB-T crowd counting methods design fusion strategies based on the assumption 
of strong modality alignment. However, this assumption rarely holds in 
practice. Rather than attempting to align mismatched modalities, we circumvent 
this challenge by using depth as a bridge modality and pre-training a 
ControlNet $f_\theta$ that captures the relationship between depth and RGB 
domains, enabling thermal-only inference. In this section, we detail the 
overall framework, the deterministic latent input design, and the initial-step 
denoising strategy.

\subsection{Overall Framework}

The overall framework of TDCount is illustrated in Figure~\ref{fig:overall}. For a given thermal image $T_i$, we first employ a backbone network to extract thermal features $F_\text{T}$. While containing important counting cues, these features lack discriminative power for high-precision counting. In parallel, we apply a depth estimation model on $T_i$ to generate the corresponding depth map $D^T_i$. The generated depth map is then fed into the pre-trained $f_\theta$, adapted with LCM LoRA~\cite{luo2023lcm}, from which we extract features $F_\text{TD}$ from the UNet's final layer before latent space conversion. These depth-derived features are used to enhance the discriminative capability of $F_\text{T}$ through a simple feature enhancement block that employs two cross-attention modules to strengthen the connections between both modalities. The enhanced features are then fed into a regression head to generate the final density map.

To adapt the generative diffusion model for the counting task and compensate for potential errors in depth estimation, we fine-tune $f_\theta$ with a task-specific LoRA that optimizes the feature representations for crowd density estimation. Additionally, we enhance the local density discrimination capability of $F_\text{T}$ by leveraging a prototype alignment loss $\mathcal{L}_\text{PA}$, formulated as:

\begin{equation}
    \mathcal{L}_\text{PA} = -\frac{1}{|\mathcal{S}|} \sum_{(j,k) \in \mathcal{S}} \log \frac{\exp(\texttt{sim}(F_\text{T}^{(j,k)}, p_{y_{j,k}}) / \kappa)}{\sum_{c=0}^{n-1} \exp(\texttt{sim}(F_\text{T}^{(j,k)}, p_c) / \kappa)},
\end{equation}

where $S$ denotes sampled spatial positions, $\text{sim}(\cdot,\cdot)$ denotes cosine similarity, $\kappa$ is a temperature factor, and $\{p_c\}_{c=0}^{n-1}$ are textual count prototypes encoded by CLIP's text encoder, where each prototype $p_c$ corresponds to a natural language description of the form ``There are $c$ persons in the area.''~\cite{ma2024clip}. This choice is further motivated by the heterogeneous nature of $F_\text{T}$ and $F_\text{TD}$. As the diffusion model is trained with CLIP text conditioning, $F_\text{TD}$ naturally embeds CLIP's semantic structure. Aligning $F_\text{T}$ with prototypes from the same CLIP text encoder therefore encourages it to learn a metric consistent with CLIP's perception of crowd counts, facilitating more effective integration with $F_\text{TD}$.

The overall training objective is:
\begin{equation}
    \mathcal{L} = \mathcal{L}_\text{Reg} + \lambda \cdot \mathcal{L}_\text{PA},
\end{equation}

where $\mathcal{L}_\text{Reg}$ is the regression loss adopted from CUT~\cite{qian2022seg} and $\lambda$ is a balancing factor.

\subsection{Deterministic Latent Input for Feature Extraction}

Latent diffusion models~(LDMs) learn a reverse Markov process that gradually denoises samples from $\mathcal{N}(\mathbf{0}, \mathbf{I})$ toward the data distribution $p(z_0)$, building rich semantic priors of the visual world that enable high-quality and diverse image synthesis. This denoising process is formalized as:
\begin{equation}
    p_\theta(z_{\tau-1}|z_\tau) = \mathcal{N}(z_{\tau-1};\, 
    \mu_\theta(z_\tau, \tau), \Sigma_\theta(z_\tau, \tau)),
\end{equation}
where $z_\tau$ denotes the noisy latent at diffusion timestep $\tau$, and $\mu_\theta(z_\tau, \tau)$ and $\Sigma_\theta(z_\tau, \tau)$ are parameterized by a UNet. ControlNet extends this framework by incorporating additional image-based conditioning signals into the reverse process, enabling the learned priors to be guided by complementary visual inputs. In our framework, we leverage a ControlNet $f_\theta$ to extract 
$F_\text{TD}$ by conditioning the denoising process on the thermal 
depth map $D^T_i$:
\begin{equation}
    F_{\text{TD}} = f_\theta(z_\tau, \tau, D^T_i).
\end{equation}
While image generation exploits the stochasticity of sampled trajectories to produce diverse outputs, our feature extraction demands consistent and reproducible representations. Specifically, if $z_{\mathcal{T}}$ were independently sampled at each training iteration, the same input $D^T_i$ would induce different trajectories through the latent space, yielding inconsistent $F_\text{TD}$ across iterations. This trajectory inconsistency 
introduces a many-to-one mapping, where the same $D^T_i$ yields 
arbitrarily different $F_\text{TD}$ across iterations. In the context of crowd counting, the regression head is required to produce a deterministic density map from these inconsistent features, forcing the model to simultaneously fit contradictory optimization targets for identical inputs. This renders the loss landscape pathological, as the same input carries different gradient signals at each iteration, making convergence impossible.

To address this, we fix $z_{\mathcal{T}}$ as a shared constant sampled once from $\mathcal{N}(\mathbf{0}, \mathbf{I})$ and held fixed across both training and inference. Under this formulation, $D^T_i$ becomes the sole source of variation in $F_\text{TD}$, ensuring a stationary and well-defined optimization target.

\subsection{The Rationale for Initial-Step Denoising Features}

To understand the rationale behind using only the first denoising step, we briefly introduce the development from LDM to LCM and explain our design choice.

Traditional LDMs require tens to hundreds of denoising steps to develop meaningful semantic representations from $z_{\mathcal{T}}$~\cite{baranchuk2022labelefficient}, making them computationally prohibitive as feature extractors. LCMs address this by enforcing a consistency constraint during training:
\begin{equation}
    f_\theta(z_\tau, \tau) = f_\theta(z_{\tau'}, \tau') \equiv z_0, 
    \quad \forall \tau, \tau' \in [0, \mathcal{T}],
\end{equation}
which ensures that any noisy latent directly maps to the same clean output $z_0$, enabling meaningful feature extraction in a single step. Starting from a fixed $z_{\mathcal{T}}$, LCM performs an initial denoising step to obtain a clean estimate $\hat{z}_0$:
\begin{equation}
    \hat{z}_0 = f_\theta(z_{\mathcal{T}}, \tau, D^T_i).
\end{equation}
While LCM also supports multi-step inference to further refine $\hat{z}_0$ and improve visual generation quality, we argue that this iterative refinement is fundamentally at odds with our feature extraction objective. In multi-step inference, LCM re-injects noise into $\hat{z}_0$ to obtain an intermediate noisy latent $z_{\tau_n}$ before performing the next denoising step:
\begin{equation}
    z_{\tau_n} = \alpha(\tau_n)\hat{z}_0 + \sigma(\tau_n)\epsilon, 
    \quad \epsilon \sim \mathcal{N}(\mathbf{0}, \mathbf{I}),
\end{equation}
where $\alpha(\tau_n)$ and $\sigma(\tau_n)$ are noise schedule parameters.The first denoising step already introduces errors due to the domain shift between thermal-derived depth maps and the RGB domain the model was trained on. Crucially, subsequent steps do not correct these structural errors; instead, they progressively commit to the visual interpretation established in the first step, solidifying the initial errors while optimizing for perceptual quality rather than structural fidelity to $D^T_i$. As a result, 
the extracted features become increasingly decoupled from the conditioning signal $D^T_i$ with each additional step. This is consistent with the error accumulation bound established by Kim et al.~\cite{kim2023consistency}:
\begin{equation}
    d_{\textnormal{TV}}(p_{\textnormal{data}},q_{\theta^\ast},N)=
    \mathcal{O}\left(\sum_{n=0}^{N-1} \sqrt{\tau_{n}}\right),
    \label{error}
\end{equation}
where sequence of timesteps $\mathcal{T}=\tau_0>\tau_1 > \cdots > 
\tau_{N-1}$, data distribution $p_{\textnormal{data}}$ and output 
distribution of LCM $q_{\theta^\ast}$. $d_{TV}(p_{\textnormal{data}}, q_{\theta^\ast})$ is total variance distance equals to 
$\frac{1}{2}\int|p_{\textnormal{data}}(x)-q_{\theta^\ast}(x)|dx$, 
$\mathcal{O}$ describes the asymptotic upper bound, meaning there exist constants $C>0$ such that $d_{TV}(p_{\textnormal{data}},q_{\theta^\ast})
\leq C\cdot\sum_{n=0}^{N-1}\sqrt{\tau_n}$ for all large enough values of $\{\tau_n\}_{n=0}^{N-1}$.

Equation~\ref{error} reveals that the total variation distance grows monotonically with the number of steps $N$, confirming that features extracted after more denoising steps diverge increasingly from the true data distribution. For crowd counting, where accurate structural correspondence between $F_\text{TD}$ and $D^T_i$ is essential for reliable density estimation, this progressive divergence directly degrades feature quality. The single-step estimate $\hat{z}_0$ therefore represents the feature most faithful to the structural content of $D^T_i$, and is the optimal choice for our counting task. This is empirically validated in Table~\ref{tab:steps}, where performance degrades monotonically as the number of denoising steps increases.

\section{Experiments}
\begin{table*}[t]
    \centering
    \caption{Comparison with the state-of-the-art methods on the RGBT-CC dataset. The \textbf{best} and \underline{second-best} results are highlighted in bold and underlined, respectively.}
    \resizebox{1.0\linewidth}{!}{
    \begin{tabular}{l|c|c|ccccc}
    \hline
    Method  & Venue & Type & GAME(0)$\downarrow$ & GAME(1)$\downarrow$ & GAME(2)$\downarrow$ & GAME(3)$\downarrow$ & RMSE$\downarrow$ \\\hline
    %UCNet~\cite{zhang2021uncertainty} & CVPR2020 & RGB-T & 33.96 & 42.42 & 53.06 & 65.07 &  56.31\\
    %HDFNet~\cite{pang2020hierarchical} & ECCV2020 & RGB-T & 22.36 & 27.79 & 33.68 & 42.48 & 33.93\\
    BL+IADM~\cite{liu2021cross} & CVPR2021 & RGB-T & 15.61 & 19.95 & 24.69 & 32.89 & 28.18\\
    CSCA ~\cite{zhang2022spatio} & ACCV2022 & RGB-T & 14.32 & 18.91 & 23.81 & 32.47 & 26.01\\
    BL+MAT~\cite{wu2022multimodal} & ICME2022 & RGB-T & 13.61 & 18.08 & 22.79 & 31.35 & 24.48\\
    DEFNet~\cite{zhou2022defnet}& TITS2022 & RGB-T & 11.90 & 16.08 &  20.19 & 27.27 & 21.09\\
    MSDTrans~\cite{liu2023rgb}  & BMVC2022 & RGB-T & 10.90 & 14.81 & 19.02 &  26.14 & 18.79\\
    TAFNet ~\cite{tang2022tafnet}& ISCAS2022 & RGB-T & 12.38 & 16.98 & 21.86 &  30.19 & 22.45\\
    R2T ~\cite{li2022learning} & KBS2022 & RGB-T & 11.63 & 16.70 & 22.12 & 32.32 & 21.28\\
    MC$^3$Net~\cite{zhou2024mc3} & TITS2023 & RGB-T & 11.47 & 15.06 & 19.40 &  27.95 & 20.59 \\
    VPMFNet~\cite{mu2024visual}   & IoT2024 & RGB-T & 10.99 & 15.17 & 20.07 & 28.03 & 19.61\\
    BGDFNet~\cite{xie2024bgdfnet}   & TIM2024 & RGB-T & 11.00 & 15.04 & 19.86 & 29.72 &  19.05\\
    C4-MIM~\cite{guo2024consistency}    & CIS2024 & RGB-T & 11.08 & 14.85 &  19.05 & 24.94 & 19.71\\
    CAGNet~\cite{yang2024cagnet}    & ESWA2024 & RGB-T & 11.06 & 14.73 &  18.94 & 25.76 & 17.83\\
    ME~\cite{Wang_2024_BMVC} & BMVC2024 & RGB-T & 11.23 & 14.98 & 18.91 & 26.54 & 19.85\\
    MJPNet-T~\cite{zhou2024mjpnet}  & IoT2024 & RGB-T &  11.56 & 16.36 &  20.95 & 28.91 & \underline{17.83}\\
    BM~\cite{meng2024multi} & ECCV2024 & RGB-T & \textbf{10.19} & \textbf{13.61} & \textbf{17.65} & \textbf{23.64} & \textbf{17.32} \\
    %GLFNet~\cite{hu2025glfnet}  & TIM2025 & RGB-T & 10.87  & 13.93 & 18.25  & 23.79 & 18.64\\
    RGBT-Booster~\cite{mu2025rgbt} & IOT2025 & RGB-T &  10.91 & 14.65 & 18.75  & 25.73 & 19.96 \\
    MIST-Net~\cite{mu2025misf} & TMM2025 & RGB-T & 10.90  & 14.87 & 19.65  & 29.18 & 19.42 \\
    CSCA~\cite{zhang2025memory} & PR2025 & RGB-T & 13.50  & 18.63 & 23.59  & 31.59 & 24.83\\
    MSPNet~\cite{liu2025modal}  & TCE2025 & RGB-T & 12.20 &16.50 &20.51&27.84& 21.49 \\
    CMFX~\cite{duan2025cmfx}  & NN2025 & RGB-T & 11.25&	15.33	&19.62	&26.14&	19.38\\\hline
    Ours      & - & T & \underline{10.62} & \underline{14.04} & \underline{17.69} & \textbf{23.64} & 19.57\\\hline
    \end{tabular}}
    \label{tab:rgbtcc}
\end{table*}
\begin{table*}[!ht]
    \centering
    \caption{Comparison with the state-of-the-art methods on the DroneRGBT dataset. The \textbf{best} and \underline{second-best} results are highlighted in bold and underlined, respectively..}
    \resizebox{1.0\linewidth}{!}{
    \begin{tabular}{l|c|c|ccccc}
    \hline
    Method  & Venue & Type & GAME(0)$\downarrow$ & GAME(1)$\downarrow$ & GAME(2)$\downarrow$ & GAME(3)$\downarrow$ & RMSE$\downarrow$ \\\hline
    BL+IADM~\cite{liu2021cross} & CVPR2021 & RGB-T & 9.74 & 12.60 & 16.49 & 21.82 & 15.37\\
    CSCA~\cite{zhang2022spatio} & ACCV2022 & RGB-T & 9.47 & 11.64 & 14.76 & 19.71 & 15.28\\
    BL+MAT~\cite{wu2022multimodal} & ICME2022 & RGB-T & 6.61 & 8.16 &\underline{10.09} &16.62 & 9.39\\
    MC$^3$Net~\cite{zhou2024mc3} & TITS2023 & RGB-T & 7.20 & 8.90 & 11.73 &  16.81 & 11.41 \\
    MJPNet-T~\cite{zhou2024mjpnet}  & IoT2024 & RGB-T & 6.02 & \underline{7.72} &  10.32 & 14.39 & 9.62\\
    BM~\cite{meng2024multi} & ECCV2024 & RGB-T &6.20 & -&-&-&10.40\\
    CAGNet~\cite{yang2024cagnet}    & ESWA2024 & RGB-T & 6.48 & 8.33 &  10.86 & \underline{14.29} & 10.30\\
    RGBT-Booster~\cite{mu2025rgbt} & IOT2025 & RGB-T &  \underline{5.91} & 7.80 & 10.45  & 14.48 & \underline{9.48} \\
    CMFX~\cite{duan2025cmfx}  & NN2025 & RGB-T & 6.75&8.88&11.87&14.69&11.05\\
    \hline Ours      & - & T & \textbf{5.65} & \textbf{7.38} & \textbf{9.69} & \textbf{13.30} & \textbf{9.09}\\\hline
    \end{tabular}}
    \label{tab:dronergbt}
\end{table*}
\subsection{Datasets and Evaluation metrics}
We conduct experiments on two widely-used RGB-T crowd counting datasets:
\begin{itemize}
    \item \textbf{RGBT-CC}~\cite{liu2021cross} is the largest RGB-T crowd counting dataset, containing 2,030 RGB-thermal image pairs with a resolution of 640×480 and 138,389 annotated pedestrians captured in diverse scenarios including malls, streets, and train stations. Among the 2,030 images, 1,013 images are captured under light conditions and 1,017 images are captured under dark conditions.
    \item \textbf{DroneRGBT}~\cite{Peng_2020_ACCV} is a drone-based RGB-T crowd counting dataset containing 3,607 registered RGB-thermal image pairs with a resolution of 640×512 and 175,698 annotated pedestrians. It contains diverse scenarios including campus, street, park, parking lot, playground, and plaza, covering different attributes such as height, illumination, and density. The dataset includes approximately 1,600 images captured under dusk conditions, 1,300 under light conditions, and 900 under dark conditions.
    \end{itemize}

We adopt the Grid Average Mean Absolute Error (GAME)~\cite{guerrero2015extremely} and the Root Mean Square Error (RMSE) to evaluate the performance. GAME is defined as:
\begin{equation}
    \text{GAME}(L) = \frac{1}{N}\sum_{i=1}^{N}\sum_{l=1}^{4^L}\left| p_i^l - g_i^l \right|,
\end{equation}
where $N$ is the number of test images, $p_i^l$ and $g_i^l$ denote the predicted and ground-truth counts in the $l$-th region of the $i$-th image. RMSE is defined as:
\begin{equation}
    \text{RMSE} = \sqrt{\frac{1}{N}\sum_{i=1}^{N}\left( p_i - g_i \right)^2},
\end{equation}
where $p_i$ and $g_i$ denote the total predicted and ground-truth counts for the $i$-th image.

\subsection{Experimental Setting}
\subsubsection{Depth-to-RGB ControlNet Training}
We fine-tune the pre-trained depth-to-RGB ControlNet on each benchmark using the training set, where depth maps are generated from RGB images using Depth Anything V2 (Large)~\cite{yang2024depth}. The model is optimized using AdamW with Pseudo Huber loss~\cite{song2023improved} over 20,000 training steps. We employ a batch size of 8 with gradient accumulation steps of 8, and images are cropped to 384×384 resolution during training. The text prompt is fixed as ``a photo of a crowd of people" with a 50\% dropout rate to encourage unconditional generation capabilities.
\begin{figure*}[t]
    \centering    \includegraphics[width=0.95\linewidth]{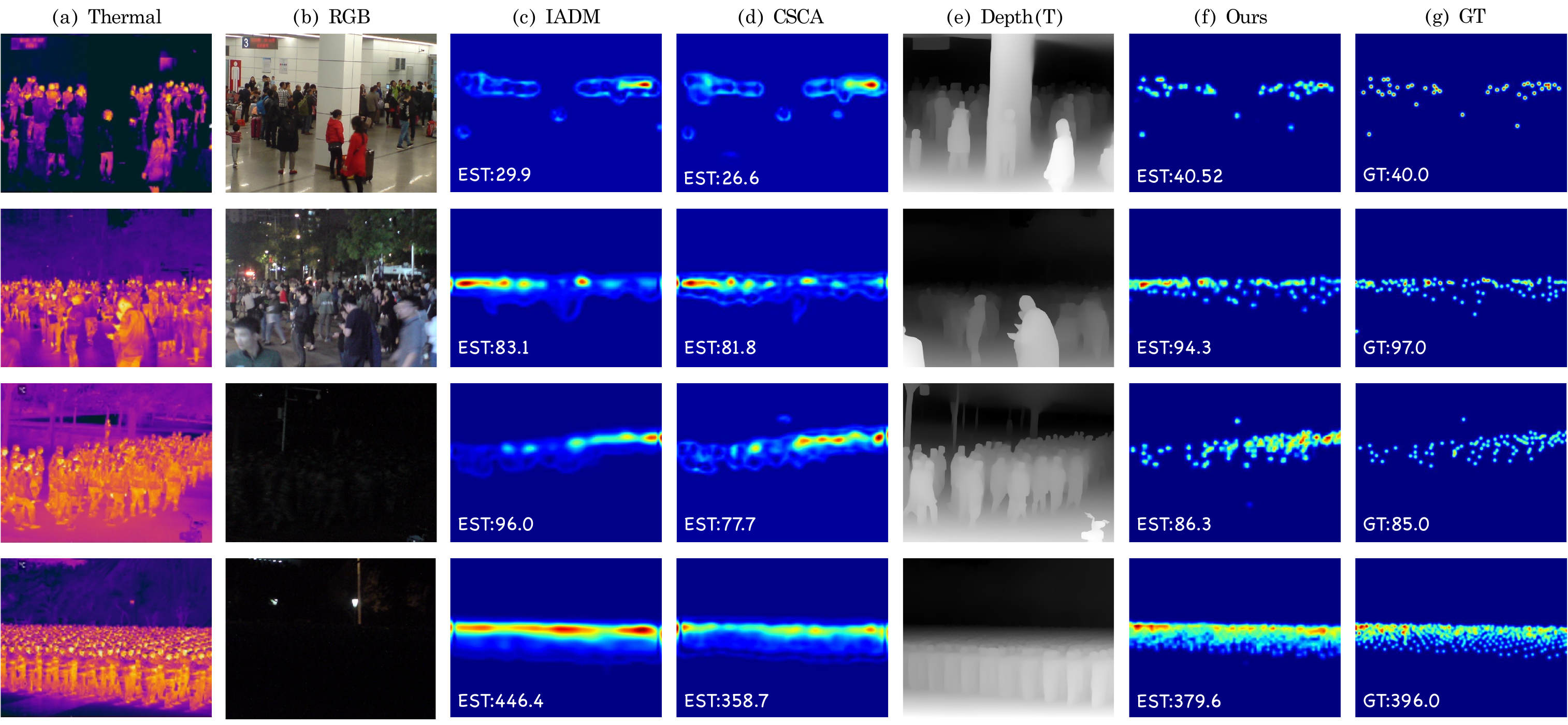}
    \caption{Visualization results of our thermal-only method versus traditional RGB-T approaches. Our method produces accurate density maps using only thermal images at inference, avoiding privacy concerns associated with RGB data while outperforming RGB-T fusion methods.}
    \label{fig:vis}
\end{figure*}

\subsubsection{TDCount Training}
We first generate ground-truth density maps for both datasets by using the fixed Gaussian kernel with a kernel size of 4. For depth extraction, we use the same depth estimation model~\cite{yang2024depth} to process thermal images, maintaining consistency with the ControlNet training stage. While the depth model is trained on RGB images, the domain shift to thermal manifests as a systematic transformation rather than random noise. The depth model consistently maps thermal intensity patterns to structural information, albeit with a characteristic bias. Crucially, this bias is consistent across all images and is thus learnable. In fact, learning to accommodate and leverage this systematic transformation is an integral part of our training framework.

We use VGG-19 as the backbone to extract features from thermal images, following common practice in the field. It is initialized with the weights that has been pretrained on the ImageNet. The regression head consists of four consecutive convolutional layers, each followed by ReLU activation, to produce the final density predictions. The counting network is trained with a batch size of 1 using AdamW optimizer. The learning rate is set to $10^{-4}$ with weight decay of $10^{-4}$. We apply random cropping for each training image. The crop size is 384\(\times\) 384. The number of textual count prototypes is set to 6 for RGBT-CC and 4 for DroneRGBT, respectively. The balancing factor $\lambda$ is set as 1 in all experiments. The textual prompt ``a photo of a crowd of people" is used with 50\% dropout during training, while being consistently provided during inference. The max training epoch is set to 500.

\subsection{Results}
We benchmark the performance of our method against previous state-of-the-art RGB-T crowd counting approaches on datasets described above. Experimental results on RGBT-CC and DroneRGBT are present in Table~\ref{tab:rgbtcc} and Table~\ref{tab:dronergbt}, respectively. It is clear that the proposed thermal-only TDCount achieves highly competitive performance across both benchmark datasets, remarkably so given that it requires only a single modality at inference while competing methods rely on paired RGB-T inputs. On the RGBT-CC dataset, TDCount shows significant improvement on different evaluation metrics. For example, compared to MC$^3$Net, our method improves achievement of 0.85 and 1.02 on GAME(0) and RMSE, respectively. More impressively, on the DroneRGBT dataset, TDCount establishes new state-of-the-art performance across all evaluation metrics. Compared to the previous best method RGBT-Booster~\cite{mu2025rgbt}, our method reduces GAME(0), GAME(1), GAME(2), GAME(3), and RMSE by 0.26, 0.42, 0.76, 1.18 and 0.39, respectively. 

Furthermore, we provide visual comparison results with previous state-of-the-art methods in Figure~\ref{fig:vis} to demonstrate the quality of our generated density maps. Notably, our thermal-only approach outperforms RGB-T fusion methods even in well-lit conditions. This suggests that a sufficiently expressive thermal representation, when augmented with diffusion-derived structural priors, can capture most of the discriminative information needed for accurate crowd counting, making the additional RGB modality less critical than commonly assumed.
\begin{table*}[t]
    \centering
    \caption{The ablation study on effectiveness of ControlNet-based Depth Processing.}
    \begin{tabular}{l|ccccc}
    \hline
    & GAME(0) &GAME(1) &GAME(2) & GAME(3)& RMSE\\\hline
    Thermal Only (Baseline) & 17.05 & 18.96 & 21.54 & 26.45 & 31.07 \\
    Thermal Only (ViT-B~\cite{dosovitskiy2020vit})& 15.83 & 18.29 & 23.43 & 32.09 & 25.51 \\
    Thermal Only (PvT-L~\cite{wang2021pyramid})& 15.30  & 18.32 & 22.76 & 30.09 &24.19  \\
    Thermal Only (ConvNeXt-B~\cite{liu2022convnet})& 15.41 & 18.54 & 23.05 & 29.68 & 24.70 \\
    Thermal + Depth (w/ VGG extraction) & 14.53 & 17.07 & 19.96 & 25.35 & 27.04 \\
    Thermal + Depth (w/ ControlNet) & \textbf{10.62} & \textbf{14.04} & \textbf{17.69}& \textbf{23.64} &\textbf{19.57}\\\hline
    \end{tabular}
    \label{tab:depth}
\end{table*}
\subsection{Ablation Studies}

\subsubsection{Effectiveness of Depth Processing.}
We conduct a systematic ablation study to understand the key factors 
enabling effective thermal crowd counting. Table~\ref{tab:depth} presents our comprehensive analysis of different feature extraction strategies.

We first investigate whether thermal ambiguity can be resolved by simply 
using a stronger backbone. Starting from a thermal-only baseline with VGG-19 for feature extraction, we replace it with ViT-B~\cite{dosovitskiy2020vit}, PvT-L~\cite{wang2021pyramid}, and ConvNeXt-B~\cite{liu2022convnet}. Despite their substantially stronger capacity and diverse architectures, these models yield only marginal improvements. This indicates that the challenge 
is not encoder capacity but the inherent ambiguity of thermal modality.

\begin{figure*}[t]
    \centering
    \includegraphics[width=0.75\linewidth]{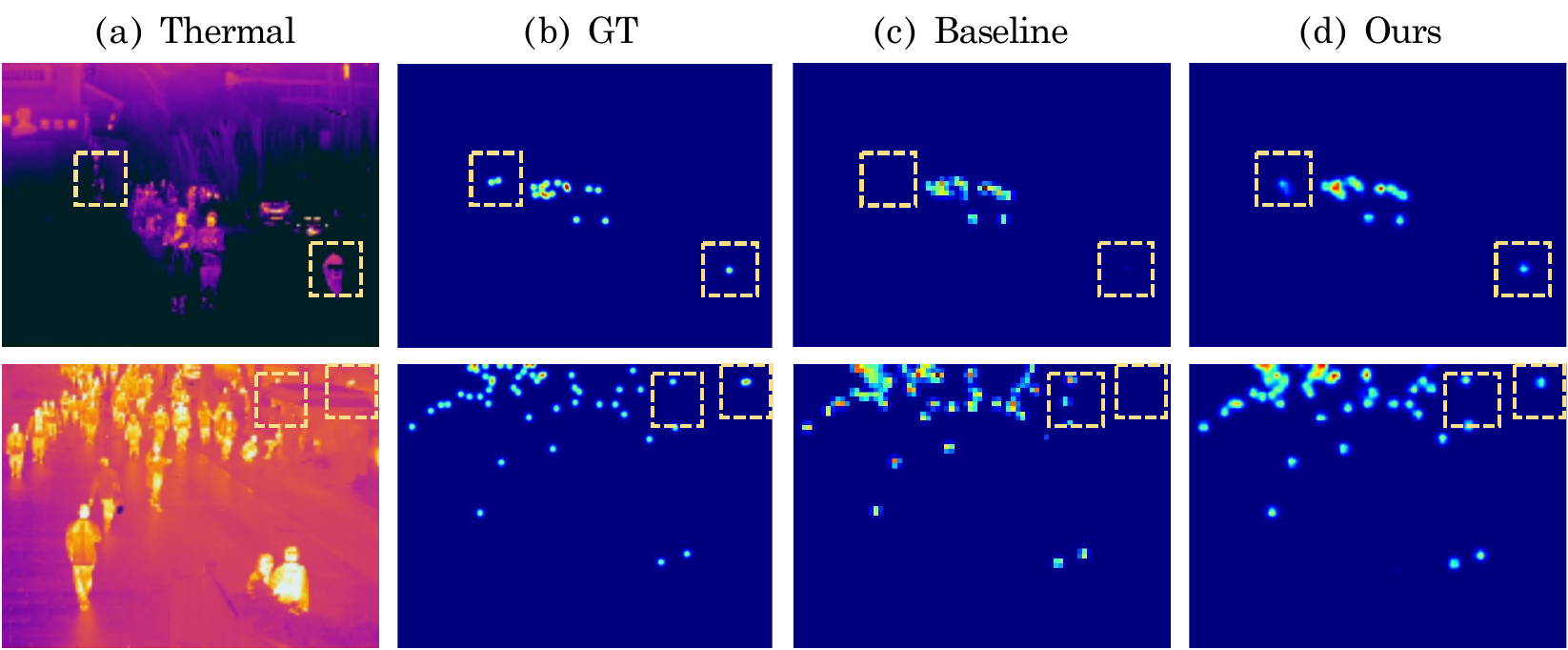}
    \caption{Qualitative comparison demonstrating the effectiveness of incorporating ControlNet-derived features. The yellow boxes highlight challenging regions where thermal signatures are ambiguous or difficult to distinguish.}
    \label{fig:ablation}
\end{figure*}

We next examine how depth information can help. We compare direct VGG-based depth feature extraction against our ControlNet-based approach. To ensure fair comparison, we apply our prototype alignment loss $\mathcal{L}_\text{PA}$ to the VGG-based depth variant as well, allowing us to isolate the effectiveness of ControlNet versus conventional depth feature extraction methods. It is clear that incorporating depth information is beneficial for thermal crowd counting, as evidenced by the improvement from thermal-only baseline achieving GAME(0) of 17.05 to VGG-based depth processing achieving GAME(0) of 14.53. Our ControlNet-based method achieves even better results with GAME(0) of 10.62, benefiting from the rich priors learned by pre-trained diffusion models that provides more discriminative features for accurate crowd density estimation. The conclusion is also verified in Figure~\ref{fig:ablation}, where our method effectively resolves ambiguous thermal signatures that challenge the thermal-only baseline.

\subsubsection{Effectiveness of the Deterministic Latent Input.}
To validate the necessity of fixing $z_{\mathcal{T}}$, we compare against a stochastic variant where $z_{\mathcal{T}}$ is re-sampled from $\mathcal{N}(\mathbf{0}, \mathbf{I})$ at each training iteration. The stochastic variant fails to converge after 500 training epochs, yielding a GAME(0) of 50.74 and RMSE of 120.65, which is even worse than the thermal-only baseline~(GAME(0): 17.05, RMSE: 31.07), indicating that the model has learned nothing meaningful from the stochastic features. By contrast, our deterministic design converges stably and achieves a GAME(0) of 10.62 and RMSE of 19.57, confirming that fixing $z_{\mathcal{T}}$ is essential for stable optimization.
% \begin{table}[tb]
%     \centering
%     \caption{The ablation on effectiveness of ControlNet-based Depth Processing. ``T'' indicates thermal, ``D'' means Depth, where ``T only'' serves as the baseline for the comparison.}
%     \begin{tabular}{l|ccc}
%     \hline
%     & T Only & T + D $_\text{(VGG)}$ & T + D $_\text{(ControlNet)}$ \\\hline
%     GAME(0) & 17.05 & 14.53 & 10.62 \\
%     GAME(1) & 18.96 & 17.07 & 14.04 \\
%     GAME(2) & 21.54 & 19.96 & 17.69 \\
%     GAME(3) & 26.45 & 25.35 & 23.64 \\
%     RMSE & 31.07 & 27.04 & 19.57 \\\hline
%     \end{tabular}
%     \label{tab:depth}
% \end{table}

\subsubsection{Influence of the Number of Denoising Steps.}
To validate our design choice of using only the first denoising step for feature extraction, we examine the influence of different denoising steps on counting performance. As shown in Table~\ref{tab:steps}, single-step denoising achieves the best results with GAME(0) of 10.62, while increasing to 2 and 3 steps leads to performance degradation with GAME(0) rising to 11.29 and 11.55, respectively. This confirms that the first denoising step captures essential structural information, additional steps may introduce cumulative errors that compromise counting accuracy. We limit our exploration to 3 steps due to computational constraints, though the declining trend already validates our single-step approach. Additionally, single-step processing offers significant computational efficiency advantages, making our method more practical for real-world deployment.
\begin{table*}[t]
    \centering
    \caption{The ablation study on the number ($n$) of denoising steps.}
    \begin{tabular}{l|ccccc}
    \hline
    & GAME(0) &GAME(1) &GAME(2) & GAME(3)& RMSE\\\hline
    $n=1$ & \textbf{10.62} & \textbf{14.04} & \textbf{17.69} & \textbf{23.64} & \textbf{19.57} \\
    $n=2$ & 11.29 & 14.43 & 17.98 & 23.72 & 20.56 \\
    $n=3$ & 11.55 & 14.97 & 18.49 & 24.19 & 20.64 \\
    $n=4$ & 11.60 & 15.28 & 18.89 & 24.85 & 22.07 \\\hline
    \end{tabular}
    \label{tab:steps}
\end{table*}
\begin{table*}[t]
    \centering
    \caption{The ablation study on effectiveness of the prototype alignment loss.}
    \begin{tabular}{l|c|ccccc}
    \hline
    & Thermal Backbone & GAME(0) &GAME(1) &GAME(2) & GAME(3)& RMSE\\\hline
    $\mathcal{L}_\text{Reg}$& \multirow{2}{*}{VGG-19}& 12.65 & 16.05 & 19.63 & 25.40 & 25.01 \\
    $\mathcal{L}_\text{Reg} + \lambda \cdot \mathcal{L}_\text{PA}$ & & \textbf{10.62} & \textbf{14.04} & \textbf{17.69} & \textbf{23.64} & \textbf{19.57}\\\hline
    $\mathcal{L}_\text{Reg}$& \multirow{2}{*}{CLIP (RN50)}&  14.87 & 17.84 & 20.91  & 26.36 & 26.71 \\
    $\mathcal{L}_\text{Reg} + \lambda \cdot \mathcal{L}_\text{PA}$ & & \textbf{13.46} & \textbf{16.15} & \textbf{19.17} & \textbf{24.67} & \textbf{25.05} \\\hline
    \end{tabular}
    \label{tab:db_loss}
\end{table*}

\subsubsection{Effectiveness of the prototype alignment loss} 
We conduct ablation studies to demonstrate the necessity of our prototype alignment loss for aligning heterogeneous feature representations. We first compare our full method with a baseline that trained with only $\mathcal{L}_\text{Reg}$ to validate the contribution of $\mathcal{L}_\text{PA}$. Subsequently, we replace the thermal backbone with a pre-trained CLIP image encoder and fine-tune the whole counting pipeline both with and without $\mathcal{L}_\text{PA}$ to verify whether this loss can also benefit when using CLIP as the thermal feature extractor. The results are given in table~\ref{tab:db_loss}. The experimental results validate the effectiveness of $\mathcal{L}_\text{PA}$ in aligning representation spaces when using VGG-19 as the thermal backbone. The prototype alignment loss improves GAME(0) from 12.65 to 10.62 and RMSE from 25.01 to 19.57, confirming the importance of bridging the heterogeneous thermal representations to the diffusion representation space. When replacing the thermal backbone with CLIP image encoder, we observe similar improvements with the $\mathcal{L}_\text{PA}$. This indicates $\mathcal{L}_\text{PA}$ helps refine globally-focused CLIP's features into more fine-grained representations suitable for counting tasks. 

\subsection{Prototype alignment Loss vs Cross Entropy Loss}

To validate the effectiveness of our prototype alignment loss, we compare $\mathcal{L}_{\text{PA}}$ against the conventional cross-entropy loss $\mathcal{L}_{\text{CE}}$ for aligning $F_\text{T}$ with $F_\text{TD}$. The results are given in table~\ref{tab:db_ce_loss}. The incorporation of cross-entropy loss $\mathcal{L}_\text{CE}$ significantly improves performance from GAME(0) of 12.65 to 11.29, confirming that adding the heterogeneous task for thermal feature learning is beneficial for the final counting accuracy. However, our proposed prototype alignment loss $\mathcal{L}_\text{PA}$ achieves further improvements, reducing GAME(0) to 10.62 and RMSE to 19.57, outperforming the cross-entropy variant by 0.67 in GAME(0) and 1.03 in RMSE. This superior performance stems from the specific design of $\mathcal{L}_\text{PA}$ that aligns thermal features with CLIP-encoded textual count prototypes through cosine similarity. Unlike generic cross-entropy supervision, our prototype alignment loss specifically targets the alignment between thermal and diffusion-derived feature spaces, leading to more effective feature integration for crowd counting.

\begin{table*}[t]
    \centering
    \caption{Further ablation study on the prototype alignment loss.}
    \begin{tabular}{l|ccccc}
    \hline
    & GAME(0) &GAME(1) &GAME(2) & GAME(3)& RMSE\\\hline
    $\mathcal{L}_\text{Reg}$& 12.65 & 16.05 & 19.63 & 25.40 & 25.01\\ 
    $\mathcal{L}_\text{Reg} + \lambda \cdot \mathcal{L}_\text{CE}$ & 11.29	& 14.46 &17.92 & 23.73 & 20.60\\\hline

    $\mathcal{L}_\text{Reg} + \lambda \cdot \mathcal{L}_\text{PA}$ & 10.62 & 14.04 & 17.69& 23.64 &19.57 \\\hline
    \end{tabular}
    \label{tab:db_ce_loss}
\end{table*}

\subsection{Ablation study on the number of textual count prototypes}
We examine how the number of textual count prototypes affects counting accuracy. The number of prototypes directly affects the granularity of the alignment. We test prototype numbers ranging from 5 to 7 to find the optimal configuration for RGBT-CC dataset. The results are present in table~\ref{tab:num_proto}, which show that $n=6$ prototypes achieves the best performance with GAME(0) of 10.62 and RMSE of 19.57. Using $n=5$ prototypes results in insufficient granularity, while $n=7$ prototypes leads to over-segmentation and slight performance degradation (GAME(0) of 10.86). We therefore adopt $n=6$ as the optimal configuration.

\begin{table*}[!htbp]
    \centering
    \caption{Ablation study on the number~($n$) of textual count prototypes.}
    \begin{tabular}{l|ccccc}
    \hline
    & GAME(0) &GAME(1) &GAME(2) & GAME(3)& RMSE\\\hline
    $n=5$ & 11.26 & 14.50 & 17.79 & 23.56 &20.73\\ 
    $n=6$ & 10.62 & 14.04 & 17.69 & 23.64 &19.57\\
    $n=7$ & 10.86 & 14.14 & 17.72 & 23.71 & 19.97 \\\hline
    \end{tabular}
    \label{tab:num_proto}
\end{table*}

\section{Discussion}
\subsection{Privacy Considerations on Training-Time RGB Usage}
A natural question is whether the use of RGB images during ControlNet fine-tuning undermines the privacy-preserving claim of our framework. We argue that the fine-tuned ControlNet does not constitute a meaningful privacy leakage channel in crowd counting scenarios, for two reasons.

First, the architectural design of latent diffusion models imposes a hard structural limit on fine-grained identity preservation. SD1.5~\cite{rombach2022high} employs a VAE encoder with an 8× spatial downsampling factor, mapping 512×512 images to 64×64 latent representations. Consequently, facial regions smaller than approximately 8×8 pixels in the original image are compressed into a single latent token or less, rendering individual identity features fundamentally unrecoverable through the denoising process. In crowd counting scenarios, where pedestrians are captured from distant viewpoints and typically occupy only a handful of pixels per head, this architectural bottleneck precludes the encoding of identity-specific information by construction.

Second, beyond architectural limits, prior work on diffusion model memorization~\cite{carlini2023extracting} has systematically shown that training data extraction is almost exclusively triggered by heavily duplicated samples (typically appearing over 100 times in the training set), while non-duplicated or out-of-distribution samples are rarely memorized even under worst-case extraction attacks. In crowd counting datasets, individual pedestrians appear only once, and we use a single generic prompt ("a photo of a crowd of people") throughout training, which provides no identity-specific conditioning signal. These conditions fall far outside the regime in which diffusion models have been shown to memorize and regenerate training content.

Taken together, the combination of architectural compression, the absence of sample duplication, and the use of a generic caption ensures that the fine-tuned ControlNet learns coarse structural priors of crowd distributions rather than identity-specific features. The privacy-preserving property of our inference pipeline is therefore not compromised by the training procedure.
\subsection{Robustness to ControlNet Checkpoint Variations}
For practical deployment, a critical question is whether our framework requires precisely tuned ControlNet models or can maintain robust performance across different post-convergence checkpoints. To investigate this practical consideration, we analyze the relationship between natural generation quality fluctuations and counting accuracy after ControlNet's sudden convergence.

The experiments are conducted on the RGBT-CC dataset. We systematically evaluate ControlNet checkpoints from 10,000 to 20,000 training steps, sampling every 500 steps. At this stage, the model has well passed the typical `sudden convergence' point that occurs around 5,000 steps and exhibits characteristic plateau behavior with natural oscillations around stable performance levels.
For each checkpoint, we measure generation quality using FID~\cite{heusel2017gans} computed on the validation set, where we sample the same regions across all images to ensure consistent evaluation. Simultaneously, we assess counting performance by integrating each checkpoint into our full pipeline and measuring GAME(0), and RMSE on the test set.

\begin{figure}[t]
    \centering
    \includegraphics[width=\linewidth]{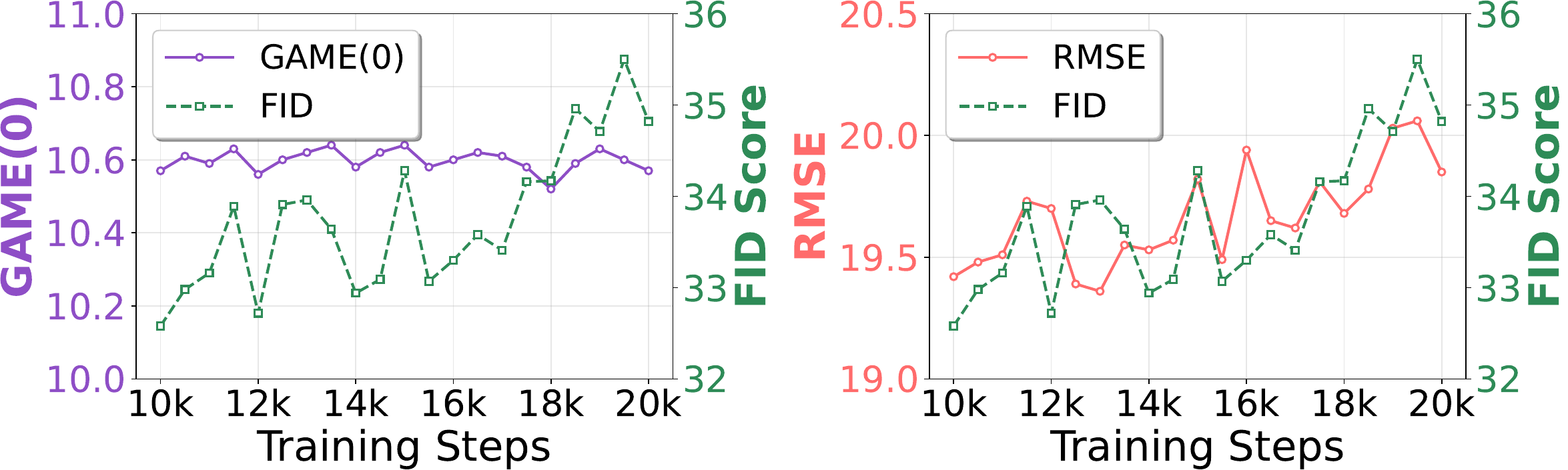}
    \caption{Post-convergence analysis demonstrating counting performance stability despite generation quality fluctuations. (a) GAME(0) and (b) RMSE show minimal variation while FID fluctuates across training steps 10k-20k.}
    \label{fig:control-qualiyu}
\end{figure}

As illustrated in Figure~\ref{fig:control-qualiyu}, FID scores exhibit fluctuations of 9.2\% throughout the 10k-20k training steps, while counting performance remains considerably more stable with GAME(0) varying only 1.1\% and RMSE varying 3.6\%. This reveals a decoupling between ControlNet's image generation quality and feature representation quality for counting tasks. The fine-grained visual details refined during post-convergence training have minimal impact on counting accuracy, as our framework primarily leverages coarse structural and semantic information rather than precise visual fidelity. Consequently, visual generation quality and counting performance show limited correlation. From a practical perspective, a reasonable ControlNet with moderate convergence is sufficient for effective crowd counting, significantly reducing computational requirements while maintaining competitive performance.

\section{Conclusion}

In this work, we propose TDCount, the first thermal-only framework for crowd counting that addresses the limitations of existing RGB-T approaches. While traditional RGB-T methods rely on continuous RGB capture at deployment time—raising privacy concerns and introducing multi-modal misalignment that degrades performance—our framework requires only thermal input at inference, with RGB used solely as a one-time, offline signal during ControlNet training on curated benchmarks. This privacy-conscious design eliminates continuous RGB capture at deployment time—the primary privacy exposure vector in real-world surveillance—substantially reducing privacy risk compared to existing RGB-T approaches. By leveraging the strong priors learned by diffusion models alongside text-guided alignment, our approach enhances thermal feature representations for more effective crowd density estimation. Our analysis further reveals that single-step LCM denoising yields features most faithful to the structural content of the depth conditioning signal, while multi-step approaches progressively decouple features from the conditioning input and accumulate errors that degrade counting accuracy. Extensive experiments show that our thermal-only approach achieves competitive performance compared to state-of-the-art dual-modality methods that require both RGB and thermal inputs at inference, demonstrating the effectiveness of our framework for privacy-conscious deployment scenarios where eliminating continuous RGB capture is a practical necessity.

% or


% \section*{Acknowledgment}


\ifCLASSOPTIONcaptionsoff
  \newpage
\fi



\bibliographystyle{IEEEtran}
\bibliography{bibtex/bib/IEEEexample}
\end{document}